\documentclass[letterpaper]{article} % DO NOT CHANGE THIS

\usepackage{aaai2026}  % DO NOT CHANGE THIS

\usepackage{times}  % DO NOT CHANGE THIS
\usepackage{helvet}  % DO NOT CHANGE THIS
\usepackage{courier}  % DO NOT CHANGE THIS
\usepackage[hyphens]{url}  % DO NOT CHANGE THIS
\usepackage{graphicx} % DO NOT CHANGE THIS
\usepackage{natbib}  % DO NOT CHANGE THIS AND DO NOT ADD ANY OPTIONS TO IT
\usepackage{caption} % DO NOT CHANGE THIS AND DO NOT ADD ANY OPTIONS TO IT
\usepackage{algorithm}
\usepackage{algorithmic}

\usepackage{newfloat}
\usepackage{listings}
\DeclareCaptionStyle{ruled}{labelfont=normalfont,labelsep=colon,strut=off} % DO NOT CHANGE THIS
\floatstyle{ruled}
\newfloat{listing}{tb}{lst}{}
\floatname{listing}{Listing}
\usepackage{booktabs}
\usepackage{array}
\usepackage{pifont}
\usepackage{xcolor}
\usepackage{amsmath}
\usepackage{placeins} 

\usepackage{pdflscape} 
\usepackage{rotating}
\usepackage{multirow}

\title{Advanced Layout Analysis Models for Docling}

\author{
    Nikolaos~Livathinos,
    Christoph~Auer,
    Ahmed~Nassar,
    Rafael~Teixeira~de~Lima,
    Maksym~Lysak,
    Brown~Ebouky,
    Cesar~Berrospi,
    Michele~Dolfi,
    Panagiotis~Vagenas,
    Matteo~Omenetti,
    Kasper~Dinkla,
    Yusik~Kim,
    Valery~Weber,
    Lucas~Morin,
    Ingmar~Meijer,
    Viktor~Kuropiatnyk,
    Tim~Strohmeyer,
    A.Said~Gurbuz,
    Peter~W.~J.~Staar
}
\affiliations{
     \tt\small \{nli,cau,ahn,mly,ceb,dol,pva,dkl,vwe,inm,vku,taa\}@zurich.ibm.com, \{rtdl,Brown.Ebouky,Matteo.Omenetti1,Yusik.Kim,Tim.Strohmeyer1,abdurrahman.said.guerbuez\}@ibm.com
  \\
    IBM Research, R\"uschlikon, Switzerland \\
}

\begin{document}

\maketitle

\begin{abstract}

This technical report documents the development of novel Layout Analysis models integrated into the Docling document-conversion pipeline.
We trained several state-of-the-art object detectors based on the RT-DETR, RT-DETRv2 and DFINE architectures on a heterogeneous corpus of 150,000 documents (both openly available and proprietary).
Post-processing steps were applied to the raw detections to make them more applicable to the document conversion task.
We evaluated the effectiveness of the layout analysis on various document benchmarks using different methodologies while also measuring the runtime performance across different environments (CPU, Nvidia and Apple GPUs).
We introduce five new document layout models achieving 20.6\% - 23.9\% mAP improvement over Docling's previous baseline, with comparable or better runtime.
Our best model, ``heron-101'', attains 78\% mAP with 28 ms/image inference time on a single NVIDIA A100 GPU.
Extensive quantitative and qualitative experiments establish best practices for training, evaluating, and deploying document-layout detectors, providing actionable guidance for the document conversion community.
All trained checkpoints, code, and documentation are released under a permissive license on HuggingFace.

\end{abstract}   
\section{Introduction}
\label{sec:intro}

The heterogeneity in the styling and representation formats of documents together with the vast amounts of information stored collectively in documents make it imperative to use specialized software that converts documents into a structured format suitable for any further data analysis.

The Document Layout Analysis task identifies document elements, classifies them according to a predefined taxonomy and locates their bounding box on the page. This is an essential part in conversion pipelines such as Docling \cite{Docling1, Docling2}, Corpus Conversion Service \cite{CCS}, MinerU \cite{MinerU}, unstructured.io \cite{unstructured_io}, Marker \cite{Marker}, Pix2Text \cite{Pix2Text}.
The same is also true for multi-stage expert models like LayoutLM \cite{Dolphin}, Dolphin \cite{Dolphin}, MonkeyOCR \cite{MonkeyOCR}, or NanonetsOCR \cite{NanonetsOCR}.

\begin{figure}[htbp]
  \centering
  \includegraphics[width=0.95\linewidth]{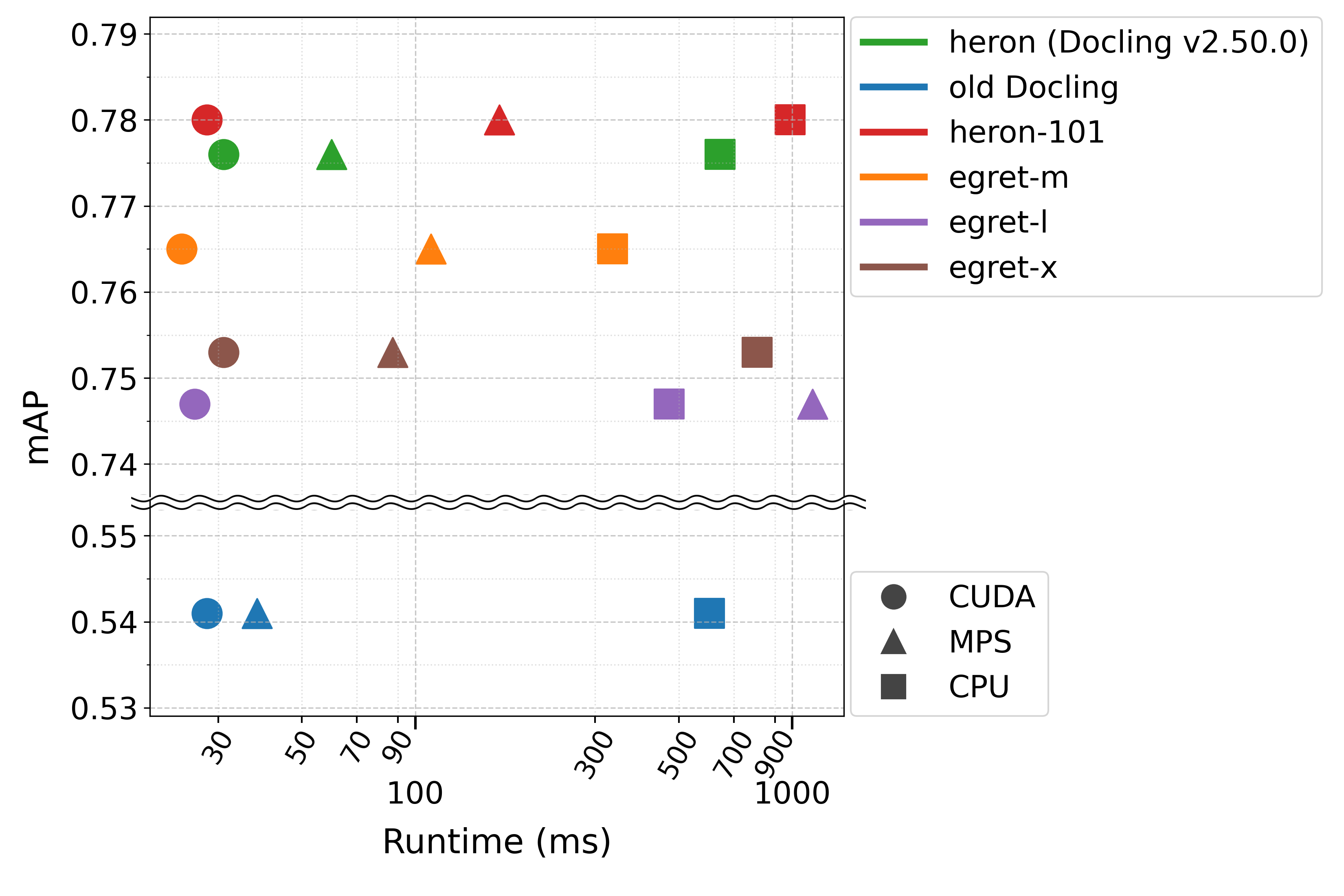}
  % \caption{Models mAP scores vs inference times. Docling's new default layout model ``heron'', provides 23.5\% mAP gain without compromising inference runtime.}
  \caption{Models mAP scores vs inference times. Docling's new default model ``heron'', delivers a 23.5\% mAP gain.}
  \label{fig:models_map_runtime}
\end{figure}

Docling is a well recognized open source\footnote{https://github.com/docling-project/docling} document conversion pipeline with a permissive license, based on top of expert AI models that we developed
and presented in the recent past \cite{SmolDocling, TableFormer, OTSL, PageModel, MolGrapher}.
Docling can recognize a wide range of document formats (PDF, MS Word, MS PowerPoint, Images, HTML, etc.) and convert them into an internal representation (\texttt{DoclingDocument}) which enables easy access to all document elements via a Python API, supports integrations with popular AI Frameworks \cite{DataPrepKit, InstructLab, BeeAI, LlamaIndex, LangChain}, exposes a Model Context Protocol interface \cite{MCP} and exports the document in structured formats like JSON, HTML, Markdown, etc.

In this technical report we present the steps we have taken to develop the family of Layout Models used by Docling.
More specifically, we outline how to:
\begin{itemize}
    \item Compile a dataset of 150k documents.
    \item Train a variety of different backbones.
    \item Apply various post-processing techniques and evaluation methodologies.
\end{itemize}

All layout models are publicly available in our Hugging Face space: \url{https://huggingface.co/ds4sd/models}{}

% Datasets:
% DocLayNet \cite{doclaynet}
% KVP10k \cite{KVP10k}
% PatCID \cite{PatCID}

\section{Data}
\label{sec:data}

We have trained the models using a diverse dataset of 150'000 single-page documents, which include a total of 2.3M document layout elements. This dataset unifies a post-processed version of the DocLayNet \cite{doclaynet} document dataset together with the proprietary dataset DocLayNet-v2 and documents from the WordScape \cite{wordscape} dataset. The unified dataset consists of 17 layout classes as presented in Table~\ref{tab:category_counts}. The annotations contain bounding boxes in COCO format and the document pages are available both as PDF files and as PNG images in 150 dpi resolution.

\begin{table}[htbp]
\setlength{\tabcolsep}{2pt} 
\centering
\begin{tabular}{lrrrr}
\hline
\textbf{Category} & \textbf{train} & \textbf{val} & \textbf{test} & \textbf{total} \\
\hline
Caption & 37,680 & 4,252 & 3,860 & 45,792 \\
Checkbox-Selected & 3,071 & 455 & 451 & 3,977 \\
Checkbox-Unselected & 45,260 & 5,827 & 6,261 & 57,348 \\
Code & 6,185 & 760 & 727 & 7,672 \\
Document Index & 1,587 & 179 & 208 & 1,974 \\
Footnote & 9,818 & 1,168 & 1,200 & 12,186 \\
Form & 12,521 & 1,625 & 1,566 & 15,712 \\
Formula & 29,101 & 2,704 & 2,923 & 34,728 \\
Key-Value Region & 20,649 & 2,665 & 2,683 & 25,997 \\
List-item & 421,845 & 47,552 & 47,426 & 516,823 \\
Page-footer & 107,761 & 11,321 & 11,304 & 130,386 \\
Page-header & 92,304 & 10,636 & 10,251 & 113,191 \\
Picture & 128,603 & 14,697 & 14,530 & 157,830 \\
Section-header & 293,021 & 34,368 & 31,212 & 358,601 \\
Table & 31,877 & 2,964 & 2,941 & 37,782 \\
Text & 1,042,044 & 118,863 & 115,479 & 1,276,386 \\
Title & 7,120 & 838 & 848 & 8,806 \\
\hline
\textbf{Total} & \textbf{2,290,447} & \textbf{260,874} & \textbf{253,870} & \textbf{2,805,191} \\
\hline
\end{tabular}
\caption{Distribution of the 17 canonical document element categories across dataset splits (train, validation, test) and total counts.}
\label{tab:category_counts}
\end{table}

% \begin{figure}[htbp]
%   \centering
%   \includegraphics[width=0.8\linewidth]{figures/GT_train_grouped_histograms.pdf}
%   \caption{Datasets distribution over the categories}
%   \label{fig:datasets_distribution}
% \end{figure}

\emph{DocLayNet} \cite{doclaynet} is a large-scale dataset for document layout analysis, offering bounding-box annotations for 80,863 unique pages across six diverse document categories. The dataset provides multilingual documents in a variety of layouts, annotated in COCO format with 150 DPI PNG images. However, a key limitation is that it includes only 11 out of the 17 canonical layout classes defined in Table~\ref{tab:category_counts}, omitting six important categories: ``Document Index'', ``Code'', ``Checkbox-Selected'', ``Checkbox-Unselected'', ``Form'', and ``Key-Value Region'' - collectively referred to as the ``delta'' classes. This omission creates a critical inconsistency for training models designed to recognize all canonical classes. Specifically, pages that do contain delta class elements are either mislabeled (e.g., a ``Form'' annotated as a ``Table'') or lack annotations altogether for those elements. As a result, these samples introduce incorrect supervision during training, which can confuse the model and degrade its ability to distinguish between similar classes. This issue becomes especially problematic in fine-grained layout tasks, where accurate class boundaries are essential for model performance.

To mitigate the negative impact of incomplete or incorrect annotations, we adopted a filtering-based approach to produce a reduced version of DocLayNet suitable for training models on all 17 canonical classes. As part of this strategy, we trained a filtering object detector—based on RT-DETRv2 on a small, curated dataset that includes all canonical categories. This detector was then applied to the full DocLayNet corpus to scan each page for the potential presence of delta class elements. Pages flagged with high likelihood of containing any of the missing classes were marked for exclusion to eliminate sources of annotation noise. To ensure high recall of delta class occurrences, we experimented with low confidence thresholds of 0.3, 0.4, and 0.5, resulting in the exclusion of 32\%, 25\%, and 20\% of the samples, respectively. Following manual inspection of the filtered outputs, we selected a threshold of 0.3, which provided the most reliable filtering performance. This process yielded a curated version of DocLayNet consisting of 22,101 training samples, 2,804 validation samples, and 1,574 test samples—ensuring that retained pages do not contain elements from the delta classes and are therefore suitable for training models targeting the complete canonical taxonomy. We call this version of DocLayNet ``canonical-DocLayNet''.

\emph{WordScape} \cite{wordscape} is an open‑source pipeline that harvests Microsoft Word documents from the Common Crawl \footnote{https://commoncrawl.org/} web corpus and converts them into a multimodal dataset suitable for training layout‑aware models. Its conversion workflow comprises three main stages: first, it extracts URLs pointing to Microsoft Word files; second, it downloads each document via HTTP requests; third, it renders every page as an image, extracts the raw text, and generates bounding‑box annotations for semantic entities such as headings and tables. We have incorporated WordScape documents from the 2013 CommonCrawl snapshot into our data mix.
However, a detailed inspection of the annotations revealed a significant semantic mismatch: WordScape’s ``Table'' label is frequently applied to entire full‑page entities which in our own classification belong to the distinct ``Form''.
Because these mislabeled instances would result to incorrect supervision during model training, we have decided to excise all WordScape documents that contain any table annotations from our dataset.

\emph{DocLayNet-v2} is an improved, proprietary version of the original dataset. It covers the full set of ``canonical'' categories listed in Table~\ref{tab:category_counts} and offers a larger and more comprehensive test split comprising 7,613 single-page documents.

%The evaluation of the models is carried out using the test splits of DocLayNet \cite{doclaynet} and DocLayNet-v2. 

\section{Object Detectors}
\label{sec:object_detectors}

In this work, we focus on Transformer-based object detectors, emphasizing fast architectures capable of real-time performance.
Because of legal constraints, we exclusively picked models released under permissive open‑source licenses.
This excludes the very popular family of YOLO \cite{YOLOv1} object detectors.

Within the DETR model family \cite{DETR}, we examined RT-DETRv1 \cite{RT-DETR} with a ResNet-50 \cite{ResNet} backbone and RT-DETRv2 \cite{RT-DETRv2} with ResNet-50 and ResNet-101 backbones.
Additionally, we assessed the DFINE detector \cite{D-FINE}, which is based on the HGNet-V2 backbone and is examined in three configurations: medium, large, and xlarge. 
% Additionally we evaluated ViTDet \citep{li2022exploringplainvisiontransformer}, an adapter for ViT-based architectures for object detection.

% Multiple implementations are available for the RT-DETRv1, RT-DETRv2, and DFINE models, including PyTorch-based versions from the original GitHub repositories\footnote{https://github.com/lyuwenyu/RT-DETR} \footnote{https://github.com/Peterande/D-FINE}, as well as implementations built on top of the Hugging Face Transformers framework \cite{transformers}.
 Multiple implementations are available for the RT-DETRv1, RT-DETRv2, and DFINE models, including PyTorch-based versions from the original GitHub repositories, as well as implementations built on top of the Hugging Face Transformers framework \cite{transformers}.
% Training with the Hugging Face Transformers implementation proved ineffective: The evaluated mAP score was very unstable and even after 2,000 steps remained below 
% In contrast, training with the PyTorch implementation provided in the original repository yielded strong results.
We trained the models with their native code implementations, then leveraged the Hugging Face utilities to convert the checkpoints from \texttt{pickle} into the \texttt{safetensors} format.
Final evaluations were then carried out using the Hugging Face inference pipeline with the converted \texttt{safetensors} checkpoints.

The RT-DETRv2-based models have been trained for 72 epochs with a learning rate of $10^{-4}$ using an AdamW[0.9, 0.999] optimizer with weight decay $10^{-4}$. 
The D-FINE models have been trained for 132, 80 and 80 epochs for the medium, large and x-large sizes, while the learning rates were $2*10^{-4}$, $2.5*10^{-4}$, $2.5*10^{-4}$ respectively. The training for the D-FINE models use an AdamW[0.9, 0.999] optimizer with weight decays $1*10^{-4}$, $1.25*10^{-4}$, $1.25*10^{-4}$. The backbones (ResNet-50, ResNet-101, HGNet-V2) have been initialized with pre-trained weights.

The training images have been augmented by a sequence of transformations. That includes randomized distortions, zoom out, horizontal flips and image resizing to 640x640.
Table~\ref{tab:model_backbones_parameters} summarizes the backbones and the number of parameters of our models.

% The ViT-Det based models have been trained following the standard procedure in \citep{li2022exploringplainvisiontransformer}. Indeed, ViT-Det \citep{li2022exploringplainvisiontransformer} is used as an adapter for ViT-based architectures to generate feature pyramid and we then use DINO \citep{zhang2022dinodetrimproveddenoising} as our decoder head. 
% The backbone, in our case DinoV2 \citep{oquab2023dinov2}, has been initialized with weights pre-trained on general-purpose images.
% The models are trained for $90k$ iterations with a learning rate of $10^{-4}$ using an AdamW[0.9, 0.999] optimizer and a weight decay of $10^{-4}$. The training images have been augmented following a sequence of augmentations such as: random flipping, resize shortest edge, and random cropping. For the implementation, we use a detrex-based repository(edge)~\footnote{https://github.com/dgcnz/edge} to run experiments on our datasets.

%%%%%%%%%%%%%%%%%%%%%%%%%%%%%%%%%%%%%%%%%%%%%%%%%%%%%%%%%%%%%%%%%%%%%%%%%%%%%%%%%%%%%%%%%%%%%%%%%%%%%
% Model backbones and number of parameters
%
\begin{table}[ht!]
\small  % Set a bit smaller font
\centering
\begin{tabular}{lll}
\toprule
Model            & Backbone         & Parameters (M) \\
\midrule
egret-m                  & DFINE-m          &  19.5 \\
egret-l                  & DFINE-l          &  31.2 \\
egret-x                  & DFINE-x          &  62.7 \\
old-docling              & RT-DETRv1-r50vd  &  42.9 \\
heron (Docling v2.50.0)  & RT-DETRv2-r50vd  &  42.9 \\
heron-101                & RT-DETRv2-r101vd &  76.7 \\
\bottomrule
\end{tabular}
\caption{Backbones and millions of parameters for Docling's layout models. Docling v2.50.0 uses ``heron'' as the default layout model.}
\label{tab:model_backbones_parameters}
\end{table}

\section{Evaluation Methods}
\label{sec:evaluation_methods}

The models were evaluated on the respective test splits of the DocLayNet \cite{doclaynet} (original and canonical) and the DocLayNet-v2 dataset. The predicted bounding boxes where evaluated once for all detections, and once for detections filtered by a minimum confidence score. Specific evaluations for Docling's output also include a set of post-processing steps.

The image input to all models were PNG images, yet the subsequent post‑processing stage also makes use of the original PDF documents by intersecting the predicted bounding box with the native PDF cells.
Each predicted label is first mapped to a minimum score; any bounding box whose confidence falls below this threshold is omitted from further analysis. The label then determines how the remaining boxes will be handled. Depending on whether it represents a ``Picture'', a wrapping document element (such as a ``Form'', ``Key Value Region'', ``Table'' or ``Document Index''), or a regular element, different processing rules apply.

For regular elements, each bounding box is matched to the best overlapping PDF cells. 
% If the ``keep‑empty'' option is disabled, any cluster that ends up with no assigned PDF cells—except for those labeled as ``Formula''—is removed. When the ``create‑orphan'' setting is enabled, orphaned PDF cells are used to create new clusters. 
The bounding boxes are then adjusted so that they exactly include their assigned PDF cells, and overlapping regular clusters are eliminated.

Special elements undergo additional treatment. Pictures that cover more than 90\% of a page are discarded outright. If a regular element overlaps with a special element, it is assigned as a child to the latter. The bounding boxes for ``Form'' and ``Key Value Region'' elements are expanded to fully enclose all their children. Overlaps involving pictures or other special types are removed.

Finally, when any group of elements overlap, only the best proposal in that group is retained; this element inherits all PDF cells from the overlapping set. The algorithm for selecting the best proposal follows a rule‑based approach that prefers certain labels, takes into account element size and confidence score, and applies different thresholds for area and confidence depending on whether the element is classified as regular, picture, or wrapper.

After applying post-processing, the results are assessed using two distinct methodologies. The first methodology relies on the \texttt{COCO-Tools} package \cite{COCO_tools}, which computes standard metrics such as Average Precision (AP) and Average Recall (AR) for each document element class at varying Intersection over Union (IoU) thresholds of 0.50, 0.75, and 0.95, as well as the mean Average Precision (mAP) averaged over a range of IoUs from 0.50 to 0.95 in increments of 0.05. Additionally, AP metrics were separately reported for small, medium, and large objects.
The second evaluation approach utilized the \texttt{docling-eval}\footnote{\url{https://github.com/docling-project/docling-eval}} package, which is part of the Docling project, providing an alternative benchmarking framework for document layout analysis.

%%%%%%%%%%%%%%%%%%%%%%%%%%%%%%%%%%%%%%%%%%%%%%%%%%%%%%%%%%%%%%%%%%%%%%%%%%%%%%%%%%%%%%%%%%%
% - First a prediction score threshold is applied to keep only predicted bounding boxes above a minimum that ranges between 0.45 and 0.50 depending on the predicted label.
% - Then all predicted scores are set to 1.0.
% - Scan the test dataset and find the intersection of labels between the prediction and the ground truth.
% - For each sample of the ground-truth and the predictions, filter out the bounding boxes to keep only the ones from the intersecting labels.
% - Skip the samples where the number of predicted bounding  boxes differs from the number of ground-truth boxes.
% - Compute the mean Average Precision for a range of IoUs from 0.50 to 0.95 in increments of 0.05.
%%%%%%%%%%%%%%%%%%%%%%%%%%%%%%%%%%%%%%%%%%%%%%%%%%%%%%%%%%%%%%%%%%%%%%%%%%%%%%%%%%%%%%%%%%%

Contrary to the \texttt{COCO-tools} which are applicable for generic object detection tasks, the \texttt{docling-eval} package evaluates the object detection in a way more targeted for documents. First, a prediction‑score threshold is applied so that only bounding boxes whose confidence lies above 0.50 are retained. Next, all remaining scores are set to 1.0. The test dataset is then scanned to identify the intersection of labels present in both the predictions and the ground truth. For each sample, only those bounding boxes whose labels belong to this intersecting set are kept; any samples where the number of predicted boxes differs from the number of ground‑truth boxes are skipped. Finally, mean Average Precision is computed over a range of Intersection‑over‑Union thresholds ranging from 0.50 to 0.95 in steps of 0.05.

\section{Results}
\label{sec:results}

The evaluation results are organized along four dimensions: (1) the evaluation dataset used, (2) the score‑threshold filter applied to predictions (3) the type of post-processing applied to the model outputs, and (4) the evaluation methodology employed (\texttt{COCO-tools} vs.\ \texttt{docling-eval}). The runtime performance measurements vary on the inference device and batch size.

%%%%%%%%%%%%%%%%%%%%%%%%%%%%%%%%%%%%%%%%%%%%%%%%%%%%%%%%%%%%%%%%%%%%%%%%%%%%%%%%%%%%%%%%%%%%%%%%%%%%%
% COCO-tools evalutions for DLNv1, DLNv2, DLNv1-canonical
\begin{table*}[ht!]
\small  % Set a bit smaller font
\centering
\begin{tabular}{cclrrrrrr}
\toprule
Dataset & Model & Post Proc. & mAP-50:95 & AP-50 & AP-75 & AP-large & AP-medium & AP-small \\

%%%%%%%%%%%%%%%%%%%%%%%%%%%%%%%%%%%%%%%%%%%%%%%%%%%%%%%%%%%%%%%%%%%%%%%%%%%%%%%%%%%%%%%%%%%%%%%%%%%%%%%%%%%%%%%%%%%%%%%%%%%%%%%%%%%%%%%%%%%%%%%%%%%%%%%
% DLNv1
\midrule
\multirow{18}{*}{DocLayNet}  & \multirow{3}{*}{egret-m}           & docling       & 0.549 & 0.659 & 0.566 & 0.515 & 0.428 & 0.455 \\
                             &                                    & direct-th50   & 0.645 & 0.791 & 0.698 & 0.670 & 0.484 & 0.419 \\
                             &                                    & direct-th0    & 0.686 & 0.848 & 0.742 & 0.705 & 0.555 & 0.484 \\
\cmidrule(r){2-9}
                             & \multirow{3}{*}{egret-l}           & docling       & 0.553 & 0.663 & 0.571 & 0.519 & 0.440 & 0.432 \\
                             &                                    & direct-th50   & 0.636 & 0.799 & 0.688 & 0.667 & 0.493 & 0.370 \\
                             &                                    & direct-th0    & 0.672 & 0.851 & 0.725 & 0.699 & 0.558 & 0.451 \\
\cmidrule(r){2-9}
                             & \multirow{3}{*}{egret-x}           & docling                   & 0.558 & 0.671 & 0.577 & 0.520 & 0.444 & 0.440 \\
                             &                                    & direct-th50   & 0.631 & 0.792 & 0.680 & 0.665 & 0.479 & 0.369 \\
                             &                                    & direct-th0    & 0.671 & 0.848 & 0.723 & 0.698 & 0.552 & 0.417 \\
\cmidrule(r){2-9}
                             & \multirow{3}{*}{old-docling}       & docling       & 0.454 & 0.558 & 0.463 & 0.442 & 0.361 & 0.397 \\
                             &                                    & direct-th50   & 0.469 & 0.677 & 0.467 & 0.502 & 0.420 & 0.241 \\
                             &                                    & direct-th0    & 0.505 & 0.733 & 0.503 & 0.539 & 0.465 & 0.314 \\
\cmidrule(r){2-9}
                             & \multirow{3}{*}{heron}   & docling                 & 0.564 & 0.674 & 0.585 & 0.521 & 0.461 & 0.457 \\
                             &                                    & direct-th50   & 0.660 & 0.805 & 0.703 & 0.678 & 0.525 & 0.440 \\
                             &                                    & direct-th0    & 0.699 & 0.859 & 0.743 & 0.712 & 0.591 & 0.519 \\
\cmidrule(r){2-9}
                             & \multirow{3}{*}{heron-101} & docling              & 0.571 & 0.683 & 0.591 & 0.529 & 0.471 & 0.484 \\
                             &                                   & direct-th50   & 0.657 & 0.799 & 0.693 & 0.672 & 0.526 & 0.447 \\
                             &                                   & direct-th0    & 0.696 & 0.851 & 0.734 & 0.707 & 0.583 & 0.478 \\

%%%%%%%%%%%%%%%%%%%%%%%%%%%%%%%%%%%%%%%%%%%%%%%%%%%%%%%%%%%%%%%%%%%%%%%%%%%%%%%%%%%%%%%%%%%%%%%%%%%%%%%%%%%%%%%%%%%%%%%%%%%%%%%%%%%%%%%%%%%%%%%%%%%%%%%
% DLNv2
\midrule
\multirow{18}{*}{DocLayNet-v2} & \multirow{3}{*}{egret-m}           & docling       & 0.232 & 0.376 & 0.197 & 0.264 & 0.135 & 0.069 \\
                               &                                    & direct-th50   & 0.673 & 0.758 & 0.726 & 0.697 & 0.530 & 0.324 \\
                               &                                    & direct-th0    & 0.725 & 0.818 & 0.785 & 0.751 & 0.581 & 0.370 \\
\cmidrule(r){2-9}
                               & \multirow{3}{*}{egret-l}           & docling       & 0.229 & 0.373 & 0.194 & 0.262 & 0.133 & 0.068 \\
                               &                                    & direct-th50   & 0.673 & 0.771 & 0.737 & 0.703 & 0.539 & 0.307 \\
                               &                                    & direct-th0    & 0.724 & 0.830 & 0.796 & 0.753 & 0.587 & 0.395 \\
\cmidrule(r){2-9}
                               & \multirow{3}{*}{egret-x}           & docling       & 0.229 & 0.371 & 0.195 & 0.261 & 0.131 & 0.072 \\
                               &                                    & direct-th50   & 0.673 & 0.762 & 0.731 & 0.702 & 0.530 & 0.318 \\
                               &                                    & direct-th0    & 0.727 & 0.825 & 0.791 & 0.755 & 0.583 & 0.383 \\
\cmidrule(r){2-9}
                               & \multirow{3}{*}{old-docling}       & docling       & 0.196 & 0.319 & 0.166 & 0.226 & 0.110 & 0.070 \\
                               &                                    & direct-th50   & 0.611 & 0.676 & 0.648 & 0.638 & 0.466 & 0.313 \\
                               &                                    & direct-th0    & 0.667 & 0.744 & 0.708 & 0.696 & 0.517 & 0.360 \\
\cmidrule(r){2-9}
                               & \multirow{3}{*}{heron}             & docling       & 0.184 & 0.298 & 0.156 & 0.208 & 0.109 & 0.055 \\
                               &                                    & direct-th50   & 0.703 & 0.779 & 0.746 & 0.724 & 0.573 & 0.346 \\
                               &                                    & direct-th0    & 0.751 & 0.832 & 0.799 & 0.771 & 0.620 & 0.416 \\
\cmidrule(r){2-9}
                               & \multirow{3}{*}{heron-101}        & docling       & 0.240 & 0.391 & 0.202 & 0.274 & 0.144 & 0.072 \\
                               &                                   & direct-th50   & 0.709 & 0.779 & 0.747 & 0.727 & 0.591 & 0.377 \\
                               &                                   & direct-th0    & 0.758 & 0.834 & 0.801 & 0.775 & 0.640 & 0.465 \\

%%%%%%%%%%%%%%%%%%%%%%%%%%%%%%%%%%%%%%%%%%%%%%%%%%%%%%%%%%%%%%%%%%%%%%%%%%%%%%%%%%%%%%%%%%%%%%%%%%%%%%%%%%%%%%%%%%%%%%%%%%%%%%%%%%%%%%%%%%%%%%%%%%%%%%%
% DLNv1-canonical
\midrule
\multirow{6}{*}{\parbox[c]{2cm}{\centering DocLayNet \\ canonical}}
                               & egret-m              & direct-th0              & 0.765             & 0.913             & 0.828             & 0.732             & 0.688             & 0.531 \\
                               & egret-l              & direct-th0              & 0.747             & 0.912             & 0.808             & 0.721             & 0.668             & 0.515 \\
                               & egret-x              & direct-th0              & 0.753             & 0.914             & 0.815             & 0.729             & 0.680             & 0.514 \\
                               & old-docling          & direct-th0              & 0.541             & 0.755             & 0.543             & 0.541             & 0.488             & 0.308 \\
                               & \underline{heron}    & \underline{direct-th0}  & \underline{0.776} & \underline{0.917} & \underline{0.826} & \underline{0.736} & \underline{0.707} & \underline{0.582} \\
                               & \textbf{heron-101}   & \textbf{direct-th0}     & \textbf{0.780}    & \textbf{0.916}    & \textbf{0.825}    & \textbf{0.738}    & \textbf{0.712}    & \textbf{0.589} \\

\bottomrule
\end{tabular}
\caption{COCO-tools evaluation on DocLayNet, DocLayNet-v2 and DocLayNet-canonical with and without post-processing. The post-processed results consider only scores above 0.5. The direct results are computed for scores above 0.5 and for all scores without any additional post-processing.}
\label{tab:coco_tools}
\end{table*}

%%%%%%%%%%%%%%%%%%%%%%%%%%%%%%%%%%%%%%%%%%%%%%%%%%%%%%%%%%%%%%%%%%%%%%%%%%%%%%%%%%%%%%%%%%%%%%%%%%%%%%%%%%%%%%%%%%%%%%%%%%%%%%%%%%%
% Discussion for the results with COCO-tools
%

Table~\ref{tab:coco_tools} presents the evaluation results using \texttt{COCO-Tools} on the original DocLayNet, on DocLayNet-v2 and on the canonical DocLayNet.
The output of each model was evaluated either directly or after applying some post-processing.
When no post-processing is applied we present the options either to evaluate on all generated predictions or to select only the ones with confidence score over 0.5.
In the case where post-processing is applied, we always keep the boxes with score over 0.5.

The evaluation with \texttt{COCO-tools} on DocLayNet shows that heron has the maximum mAP score 0.699 when no post-processing is applied and all predicted boxes are taken into account.
On the second rank, we find heron-101 with an mAP of 0.696.
All variants of post-processing damage the mAP score by more than 10\% and to a lesser extend the removal of boxes with low score.
When evaluating on DocLayNet-v2, the best mAP score is seen with heron-101 (0.758) and the second best with heron (0.751).

As already discussed, the annotations of the original DocLayNet dataset contain mismatches with our canonical classes.
These inconsistencies penalize our metrics and lower the AP scores for the examples that contain the delta classes.
To address this, we performed evaluations on the canonical version of DocLayNet with \texttt{COCO-tools} without any post-processing.
As evident in Table~\ref{tab:coco_tools} the canonical DocLayNet improves the mAP by 7.5\% to 8.4\% in comparison to the original DocLayNet.
This benchmark allows heron-101 to achieve the highest mAP score 0.780.

%%%%%%%%%%%%%%%%%%%%%%%%%%%%%%%%%%%%%%%%%%%%%%%%%%%%%%%%%%%%%%%%%%%%%%%%%%%%%%%%%%%%%%%%%%%
% TODO: Discussion for the results with docling-eval
%
The evaluation results from the \texttt{docling-eval} package are presented in Table~\ref{tab:docling-eval_DLNv1} for DocLayNet and DocLayNet-v2. In this case some type of post-processing is always applied. 
heron ranks first for both datasets while heron-101 and the egret models are slightly behind.
The substantial score gap between DocLayNet and DocLayNet‑v2 across all models reflects both the higher document layout complexity in DocLayNet‑v2 and the evaluation approach employed by \texttt{docling-eval}.

%%%%%%%%%%%%%%%%%%%%%%%%%%%%%%%%%%%%%%%%%%%%%%%%%%%%%%%%%%%%%%%%%%%%%%%%%%%%%%%%%%%%%%%%%%%%%%%%%%%%%%%%%%%%%%%%%%%%%%%%%%%%%%%%%%%
% Evaluations with docling-eval

\begin{table}[ht!]
\small  % Set a bit smaller font
\centering
\begin{tabular}{lll}
\toprule
Model            & DocLayNet & DocLayNet-v2 \\
\midrule
egret-m                & 0.59              & 0.35  \\
egret-l                & 0.59              & 0.35  \\
egret-x                & 0.60              & 0.35  \\
old-docling            & 0.47              & 0.31 \\
\textbf{heron}         & \textbf{0.61}     & \textbf{0.36} \\
\underline{heron-101}  & \underline{0.61}  & \underline{0.35} \\
\bottomrule
\end{tabular}
\caption{mAP scores on DocLayNet and DocLayNet-v2 using docling-eval with post-processing}
\label{tab:docling-eval_DLNv1}
\end{table}

% \begin{table}[ht!]
% \small  % Set a bit smaller font
% \centering
% \begin{tabular}{lll}
% \toprule
% Model            & DocLayNet & DocLayNet-v2 \\
% \midrule
% DFINE-m          & 0.59      & 0.35  \\
% DFINE-l          & 0.59      & 0.35  \\
% DFINE-x          & 0.60      & 0.35  \\
% RTDETR-v1        & 0.47      & 0.31 \\
% RTDETR-v2-r50vd  & 0.61      & 0.36 \\
% RTDETR-v2-r101vd & 0.61      & 0.35 \\
% \bottomrule
% \end{tabular}
% \caption{mAP scores on DocLayNet and DocLayNet-v2 using docling-eval with post-processing}
% \label{tab:docling-eval_DLNv1}
% \end{table}
%%%%%%%%%%%%%%%%%%%%%%%%%%%%%%%%%%%%%%%%%%%%%%%%%%%%%%%%%%%%%%%%%%%%%%%%%%%%%%%%%%%%%%%%%%%%%%%%%%%%%%%%%%%%%%%%%%%%%%%%%%%%%%%%%%%

\begin{landscape}
\begin{figure}[t]
  \noindent\makebox[\paperheight][c]{
    \hspace{-190pt}  % Ugly hack to compensate something that pushes the minipage to the right
    \begin{minipage}{0.9\paperheight}
      \centering
      \includegraphics[height=0.25\paperheight]{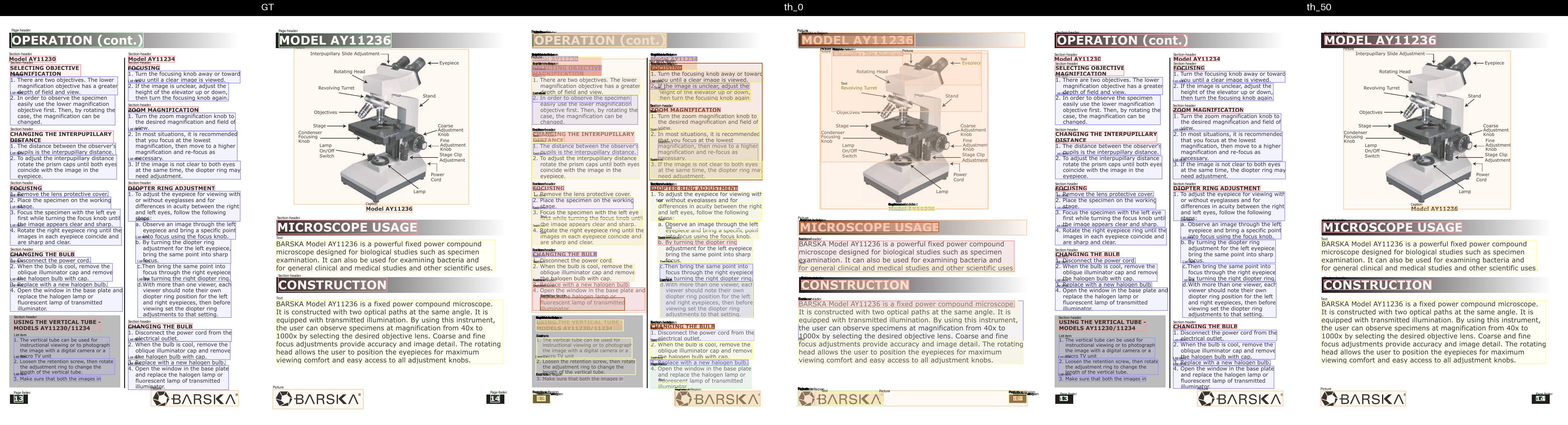}
      \includegraphics[height=0.35\paperheight]{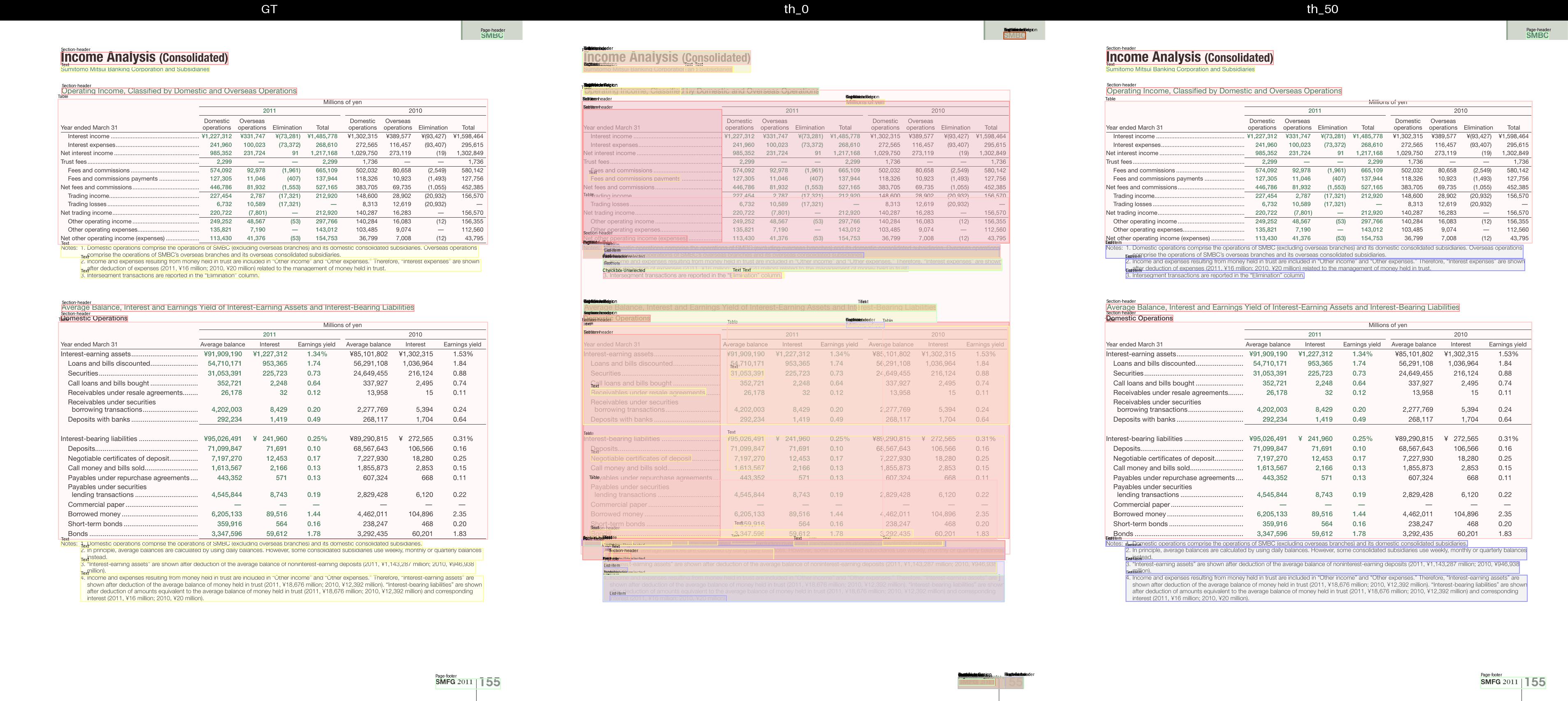}
      \caption{Predictions made by the ``heron" model. The ground truth, the predictions without score filtering and the predictions with score over 0.5 are visualized from left to right. Although the filtered predictions have lower mAP, they contain fewer overlapping bounding boxes and look closer to the ground truth.}
      \label{fig:DLNv1_preds_th0_th50}
    \end{minipage}
  }
\end{figure}
\end{landscape}

%%%%%%%%%%%%%%%%%%%%%%%%%%%%%%%%%%%%%%%%%%%%%%%%%%%%%%%%%%%%%%%%%%%%%%%%%%%%%%%%%%%%%%%%%%%
% Discussion about the applicability of mAP metric for document layout analysis
%
In addition to the quantitative evaluation presented above, we have run qualitative analysis based on visualizations of the predictions versus the ground truth.
Figure~\ref{fig:DLNv1_preds_th0_th50} depicts on the left side the ground-truth image, in the middle the raw model predictions without any post-processing and on the right side the predictions with scores over 0.50.
According to the results of Table~\ref{tab:coco_tools} the mAP score is maximized when no post-processing or score filtering is applied.
However, as it is clearly evident in Figure~\ref{fig:DLNv1_preds_th0_th50}, the raw predictions (in the middle) are very noisy with many overlapping bounding boxes.
On the other hand, the filtered predictions look much closer to the ground-truth, regardless of yielding lower mAP scores.

\begin{landscape}
\begin{figure}[t]
  \noindent\makebox[\paperheight][c]{
    \hspace{-190pt}  % Ugly hack to compensate something that pushes the minipage to the right
    \begin{minipage}{0.9\paperheight}
      \centering
      \includegraphics[height=0.30\paperheight]{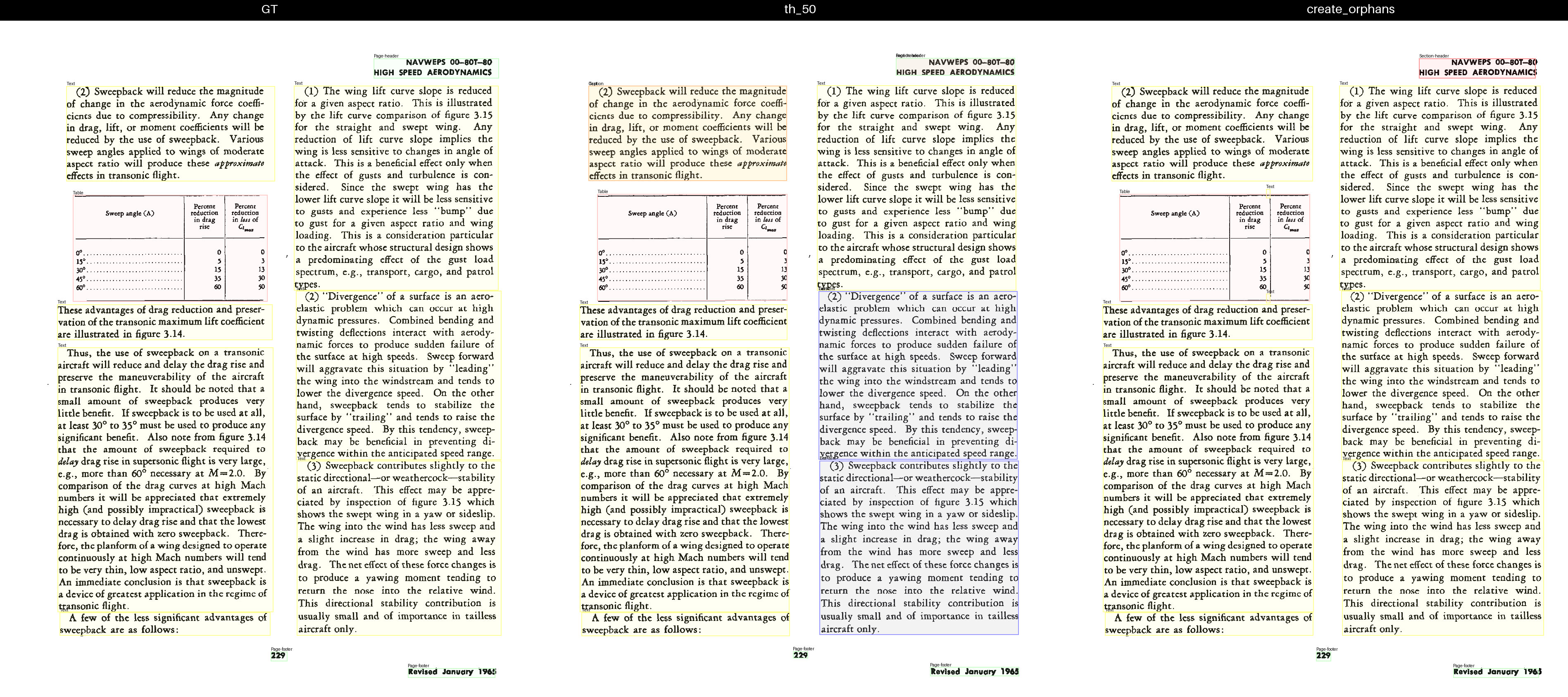}
      \includegraphics[height=0.30\paperheight]{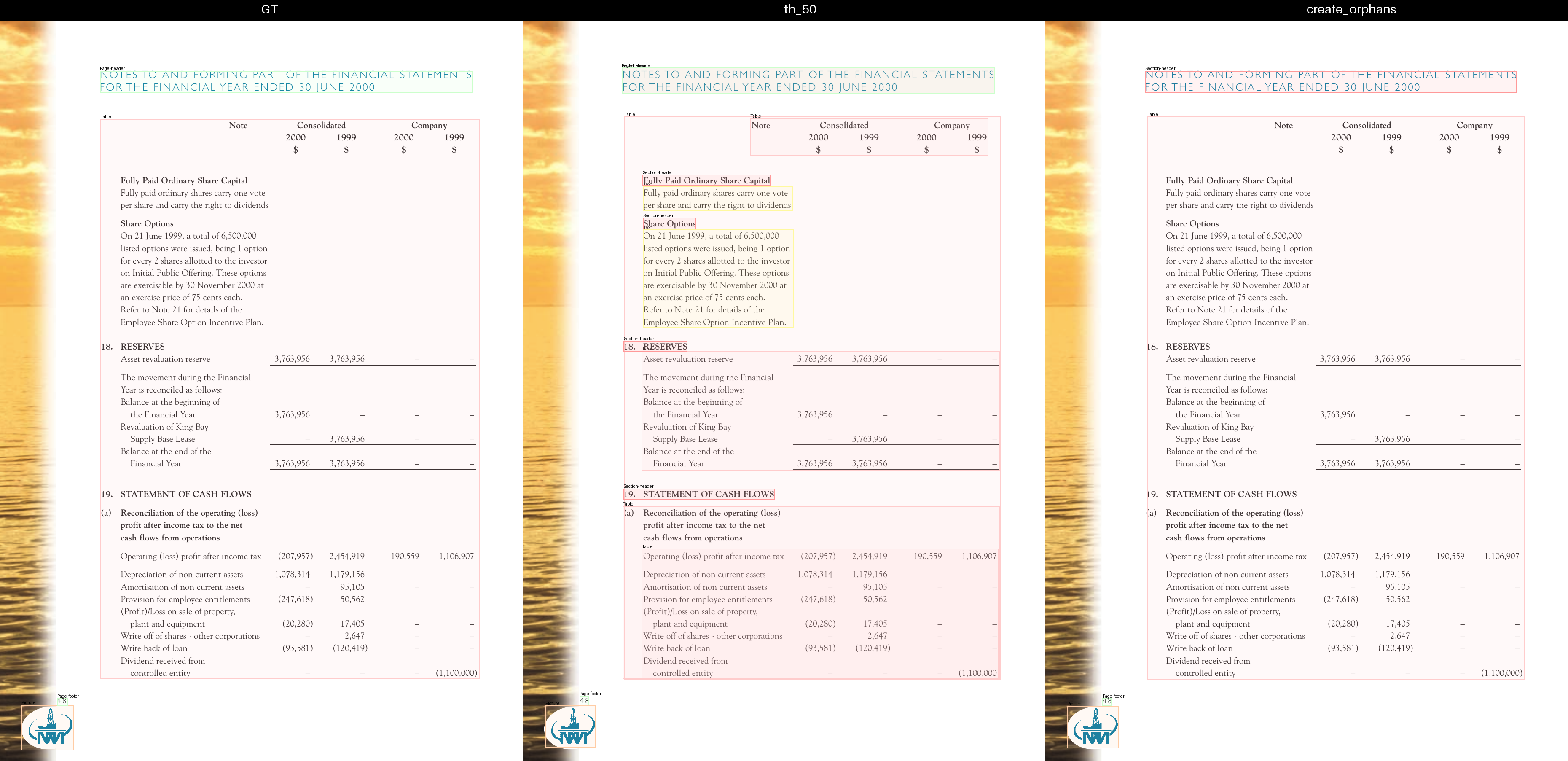}
      \caption{Predictions made by the ``heron" model with post-processing. The ground truth, the predictions with score over 0.5 and the predictions with the standard Docling post-processing are visualized from left to right. Post‑processing eliminates misidentified ``List‑item'' elements and corrects a full‑page ``Table'' from detections that were mistakenly recognized as separate document items.}
      \label{fig:DLNv1_preds_th50_orphans}
    \end{minipage}
  }
\end{figure}
\end{landscape}

Next, we compare the predictions with scores over 0.50 with the post-processed model outputs.
Figure~\ref{fig:DLNv1_preds_th50_orphans} presents the ground truth (left), the model output with scores over 0.50 (middle) and the post-processed predictions, which is the default option for Docling.
As one can see, the post-processing has improved the raw predictions by clustering together fragmented document elements and removing overlapping bounding boxes.
This has clearly improved the layout geometry of the end-result but the assigned labels do not always match the ground truth.
For example in the example of the top row of Figure~\ref{fig:DLNv1_preds_th50_orphans}, the generated cluster is assigned the label ``Picture'' but the annotation is ``Table''.

\begin{figure*}[t]
  \centering
  \includegraphics[width=0.90\textwidth]{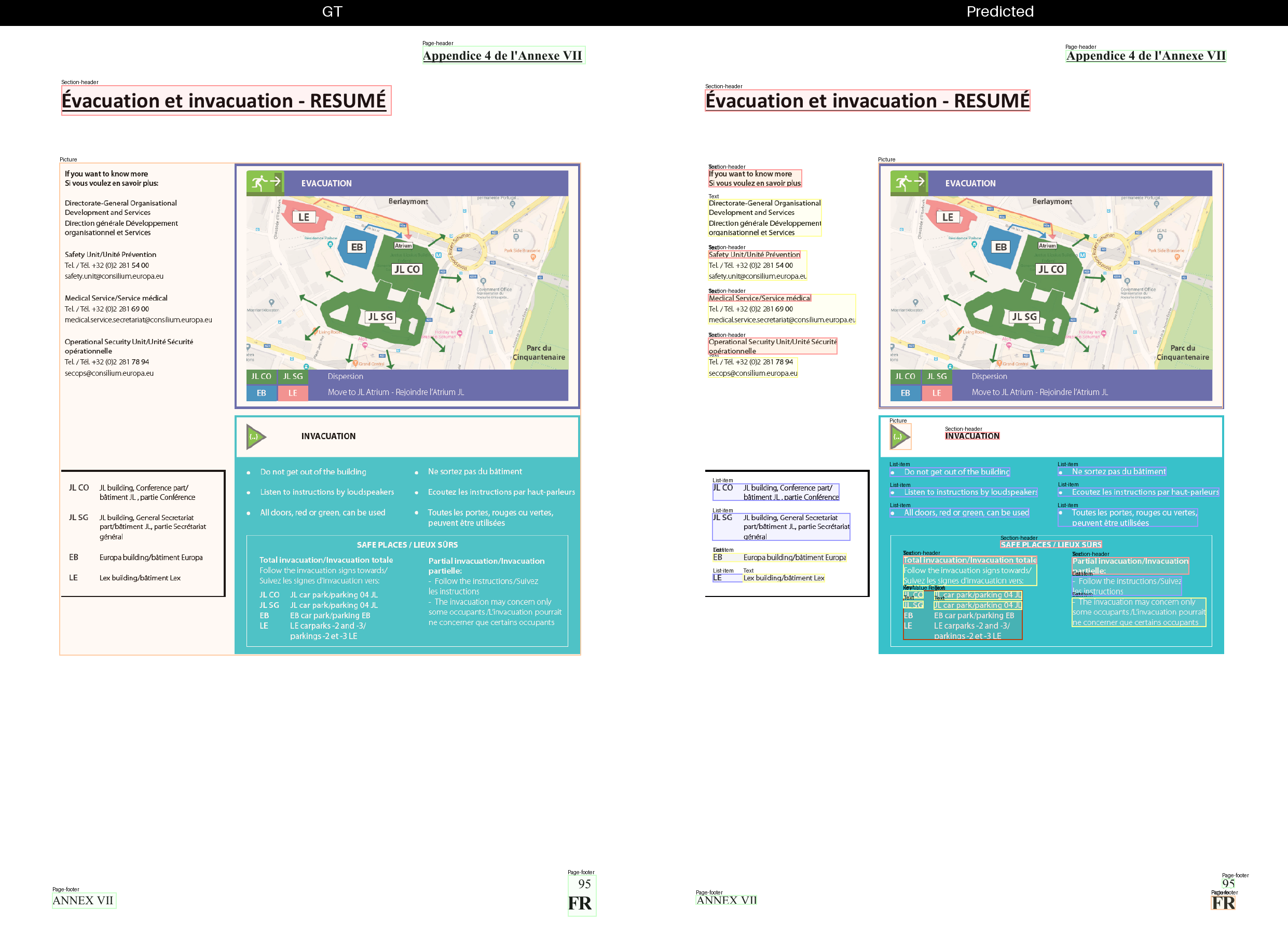}
  \hfill
  \includegraphics[width=0.90\textwidth]{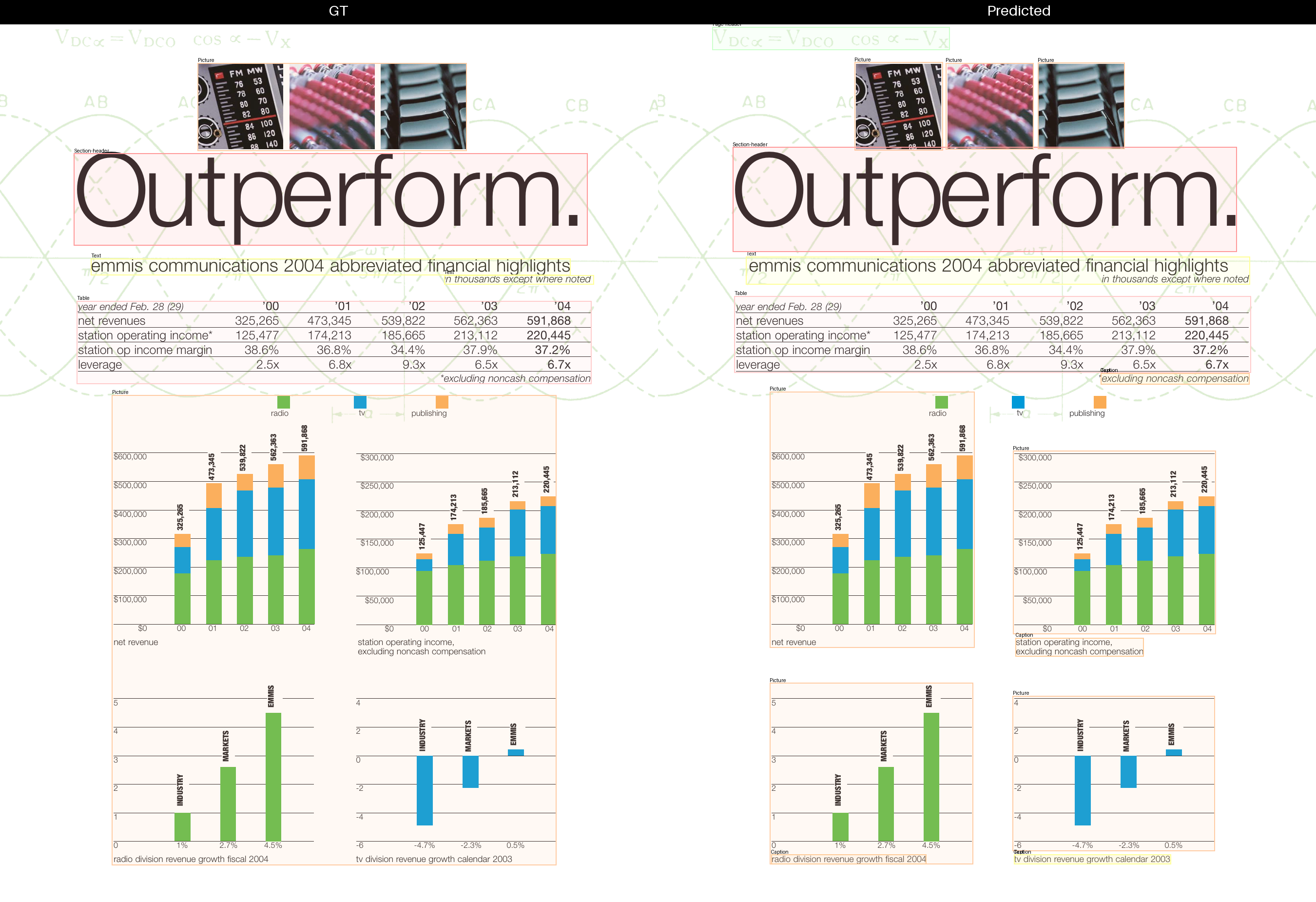}
  \caption{Sometimes alternative layout resolutions also look reasonable. Selected predictions of the ``heron'' model with score above 0.5}
  \label{fig:DLNv1_preds_alternatives}
\end{figure*}

The high complexity of document layouts often yields ambiguous annotations.
Figure~\ref{fig:DLNv1_preds_alternatives} presents cases where it is not clear if the ground truth data or the model predictions are correct or maybe \textit{both} are valid layout resolutions.
In the first example the main body of the page has been annotated as one big ``Picture'',
but the model predicts a more detailed classification where textual elements have been identified as ``Section-Header'', ``Text'' and ``List Item'' and the bounding boxes of the pictures have been reduced to cover only the visual content.
Such discrepancies suggest that alternative layouts differing both in geometry and classification may be acceptable.

The mean Average Precision (mAP) score, originally developed to evaluate general-purpose object detection,
has been widely adopted as a standard metric for document layout analysis tasks.
Based on the observations of this study and our overall experience with Docling, we conclude that mAP
may not always be a suitable metric to evaluate the layouts of documents.
% as it assumes a single correct layout resolution.

%%%%%%%%%%%%%%%%%%%%%%%%%%%%%%%%%%%%%%%%%%%%%%%%%%%%%%%%%%%%%%%%%%%%%%%%%%%%%%%%%%%%%%%%%%%
% Discussion about the runtime performance
%
Lastly we have measured the runtime performance of our models in 3 hardware configurations: a single AMD EPYC 7763 64-Core, a single Nvidia A100-80GB and an Apple MacBook M3 with MPS enabled.
We have measured the min, max, median, mean seconds needed to run the models on the test split of DocLayNet without counting the I/O operations.
The measurements have been divided over the split size to show the amortized inference time per image as shown in Table~\ref{tab:runtime}.
The lightest model, egret-m, is the fastest across all testing environments achieving 0.024 sec/image when running on A100  with batch size 200.
The runtime of our most accurate model, heron-101, is 0.028 sec/image.
On a typical server CPU with 4 threads, egret-m is three times faster than heron-101 when using batch size 32, whereas GPU performance is similar across all models
Notably, the relatively small sizes of our models pose a challenge to saturate the A100 GPU.
To mitigate this, we employed large batch sizes and parallelized data loading to the GPU using 32 CPU threads.

\section{Conclusion}
\label{sec:conclusion}

%%%%%%%%%%%%%%%%%%%%%%%%%%%%%%%%%%%%%%%%%%%%%%%%%%%%%%%%%%%%%%%%%%%%%%%%%%%%%%%%%%%%%%%%%%%
% Conclusion
%

In this technical report we presented the new family of Layout Models used by Docling.
We explained how we built a diverse dataset of 150k documents, how we trained the models and which evaluation methods were used.
Our best model, heron-101, achieves a 23.9\% improvement over Docling's prior layout model with 78\% mAP score when measured on the canonical DocLayNet dataset and is based on RT-DETRv2 architecture with a ResNet101 backbone.
Nevertheless, the inference runtime stays on par with what Docling's earlier model with 0.028 sec/image when measured on a single Nvidia A100 GPU.
The fastest model, egret-m, achieves 0.024 sec/image.
Our analysis showed that keeping only the elements with high prediction score and applying post-processing can improve the quality of the layout resolution,
however without this improvement being reflected on the mAP score.
Our observations imply that the mean Average Precision may not be the best metric for the layout analysis on documents.
You are very welcome to explore our publicly available models.

%%%%%%%%%%%%%%%%%%%%%%%%%%%%%%%%%%%%%%%%%%%%%%%%%%%%%%%%%%%%%%%%%%%%%%%%%%%%%%%%%%%%%%%%%%%%%%%%%%%%%%%%%%%%%%%%%%%%%%%%%%%%%%%
% Examples with triplets of threshold 0 and 50
%

% Original code with portrait figure of 3 examples
% \begin{figure*}[t]
%   \centering
%   \includegraphics[width=0.80\textwidth]{figures//preds_DLNv1_direct_th0_th50/alpha30/viz_1b81cf65f47456ad4faa725d1eb09879bd633af16cfe2bf8cea661b87907bfac.png}
%   \hfill
%   \includegraphics[width=0.80\textwidth]{figures//preds_DLNv1_direct_th0_th50/alpha30/viz_3a3a132491c67be1b29f0dd1f51e6af94c6f37207babc9380759ca3f04fd39b5.png}
%   \hfill
%   \includegraphics[width=0.80\textwidth]{figures//preds_DLNv1_direct_th0_th50/alpha30/viz_96ebf378e84e74d65ac69d3aabed83e883a69f6f3bf83c1d2b0c3500095301bb.png}
%   \caption{Predictions with the RT-DETRv2-r50vd. The ground truth, the predictions without score filtering and the predictions with score over 0.5 are visualised from left to right.}
%   \label{fig:DLNv1_preds_th0_th50}
% \end{figure*}

% % Just rotation of the figure, while keeping the page in portrait
% \begin{sidewaysfigure}
% \centering
% \includegraphics[width=\textwidth]{figures//preds_DLNv1_direct_th0_th50/alpha30/viz_1b81cf65f47456ad4faa725d1eb09879bd633af16cfe2bf8cea661b87907bfac.png}
% \caption{A wide figure in landscape orientation}
% \label{fig:landscape}
% \end{sidewaysfigure}

% Landscape with 2 examples
% Start landscape, add a figure (no figure *), inside the figure is a box with
% a shifted minipage that has the graphics and the caption

%%%%%%%%%%%%%%%%%%%%%%%%%%%%%%%%%%%%%%%%%%%%%%%%%%%%%%%%%%%%%%%%%%%%%%%%%%%%%%%%%%%%%%%%%%%%%%%%%%%%%%%%%%%%%%%%%%%%%%%%%%%%%%%
% Examples with triplets of threshold 50 and post-processing with orphans
%

% Original code with portrait figure of 3 examples
% \begin{figure*}[t]
%   \centering
%   \includegraphics[width=0.80\textwidth]{figures//create_orphans/alpha30/viz_0a5a8eea938173c5aebf6c4a215de7ff90377433e4417567120fc2e5109f4800.png}
%   \hfill
%   \includegraphics[width=0.80\textwidth]{figures//create_orphans/alpha30/viz_4b559a463e9a631a3a48caea9366351fd735d4e75df1405b43f4101d217d28ef.png}
%   \hfill
%   \includegraphics[width=0.80\textwidth]{figures//create_orphans/alpha30/viz_3feeb5dcdc9e8a8a96ac5243b871f7855d34a8a4f06eefe71facae4177a39c04.png}
%   \caption{Predictions with the RT-DETRv2-r50vd and post-processing. The ground truth, the predictions with score over 0.5 and the predictions with the standard Docling post-processing are visualised from left to right.}
%   \label{fig:DLNv1_preds_th50_orphans}
% \end{figure*}

% Landscape with 2 examples

%%%%%%%%%%%%%%%%%%%%%%%%%%%%%%%%%%%%%%%%%%%%%%%%%%%%%%%%%%%%%%%%%%%%%%%%%%%%%%%%%%%%%%%%%%%%%%%%%%%%%%%%%%%%%%%%%%%%%%%%%%%%%%%
% Alternative layout resolutions
%

% Original code with portrait figure of 2 examples

%%%%%%%%%%%%%%%%%%%%%%%%%%%%%%%%%%%%%%%%%%%%%%%%%%%%%%%%%%%%%%%%%%%%%%%%%%%%%%%%%%%%%%%%%%%%%%%%%%%%%%%%%%%%%%%%%%%%%%%%%%%%%%%%%%%
% Runtime evaluations
%

% 14.09.2025: Explanation about this table:
% - CPU runtimes:
%   - All experiments are done on a "AMD EPYC 7763" with 4 threads.
%   - It is imperative to run each experiment multiple times and averge to reduce measurement errors
%   - The results for CPU with batch_size = 32 were run 3 times and averaged.
%   - The results for other batch_sizes are unreliable as I did not repeat/average the experiments.
%     It would be good though to do it.
% - CUDA runtimes:
%   - I have run the experiments only once with the GPU in "exclusive" mode.
%   - I expect to have accurate numbers.
% - MPS runtimes:
%   - Skip the batch_size=1 as the measurement proves to be very unstable.
%   - For batch_sizes in [50, 100], it seems to be quite reproducible even without averaging.
%
\begin{table*}[ht!]
\small
\centering
\begin{tabular}{cclcccc}
\toprule
Device           & Batch-Size & Model                & mean  & median & min   & max   \\
\midrule
                                                     % CPU experiments run 3 times + averaged 
\multirow{6}{*}{CPU}    & \multirow{6}{*}{32}    & egret-m      &  0.334  &  0.329 &  0.075 &  0.440 \\
                        &                        & egret-l      &  0.472  &  0.463 &  0.099 &  0.682 \\
                        &                        & egret-x      &  0.808  &  0.797 &  0.170 &  1.070 \\
                        &                        & old-docling  &  0.603  &  0.596 &  0.138 &  0.783 \\
                        &                        & heron        &  0.643  &  0.639 &  0.141 &  0.826 \\
                        &                        & heron-101    &  0.988  &  0.983 &  0.216 &  1.322 \\

\midrule
\multirow{15}{*}{CUDA}  & \multirow{6}{*}{100}   & egret-m      & 0.024 & 0.023  & 0.022 & 0.043 \\
                        &                        & egret-l      & 0.027 & 0.026  & 0.025 & 0.048 \\
                        &                        & egret-x      & 0.030 & 0.029  & 0.027 & 0.047 \\
                        &                        & old-docling  & 0.025 & 0.024  & 0.024 & 0.039 \\
                        &                        & heron        & 0.030 & 0.027  & 0.026 & 0.099 \\
                        &                        & heron-101    & 0.174 & 0.168  & 0.162 & 0.259 \\
\cmidrule(r){2-7}
                        & \multirow{6}{*}{200}   & egret-m      & 0.024 & 0.023  & 0.023 & 0.034 \\
                        &                        & egret-l      & 0.026 & 0.025  & 0.025 & 0.037 \\
                        &                        & egret-x      & 0.031 & 0.028  & 0.027 & 0.087 \\
                        &                        & old-docling  & 0.028 & 0.027  & 0.026 & 0.040 \\
                        &                        & heron        & 0.031 & 0.029  & 0.027 & 0.068 \\
                        &                        & heron-101    & 0.028 & 0.028  & 0.027 & 0.038 \\
\cmidrule(r){2-7}
                        & \multirow{3}{*}{500}   & egret-m      & 0.026 & 0.026  & 0.025 & 0.026 \\
                        &                        & old-docling  & 0.029 & 0.029  & 0.027 & 0.031 \\
                        &                        & heron        & 0.026 & 0.027  & 0.026 & 0.027 \\
\midrule
\multirow{12}{*}{MPS}   & \multirow{6}{*}{50}    & egret-m      & 0.033 & 0.033 & 0.030 & 0.042 \\
                        &                        & egret-l      & 0.040 & 0.039 & 0.036 & 0.070 \\
                        &                        & egret-x      & 0.094 & 0.090 & 0.062 & 0.138 \\
                        &                        & old-docling  & 0.072 & 0.072 & 0.070 & 0.080 \\
                        &                        & heron        & 0.044 & 0.044 & 0.041 & 0.051 \\
                        &                        & heron-101    & 0.062 & 0.062 & 0.057 & 0.076 \\
\cmidrule(r){2-7}                                                                                 
                        & \multirow{6}{*}{100}   & egret-m      & 0.038 & 0.034 & 0.033 & 0.078 \\
                        &                        & egret-l      & 0.110 & 0.108 & 0.049 & 0.170 \\
                        &                        & egret-x      & 1.131 & 1.216 & 0.069 & 1.381 \\
                        &                        & old-docling  & 0.087 & 0.083 & 0.072 & 0.118 \\
                        &                        & heron        & 0.060 & 0.053 & 0.041 & 0.091 \\
                        &                        & heron-101    & 0.167 & 0.150 & 0.071 & 0.376 \\
\bottomrule
\end{tabular}

\caption{Inference runtime per image in seconds for various devices and batch sizes. CPU=AMD EPYC 7763 (4 threads). GPU=A100 80GB. MPS=M3 Max, 40 cores, 64GB.}
% All predictions are for scores more than 0.5
\label{tab:runtime}
\end{table*}

\clearpage
% \bibliography{references}

\end{document}